# Chapter 10
# An Immune-Inspired Approach to Anomaly Detection

Jamie Twycross, University of Nottingham, UK
Uwe Aickelin, University of Nottingham, UK

## ABSTRACT

The immune system provides a rich metaphor for computer security: anomaly detection that works in nature should work for machines. However, early artificial immune system approaches for computer security had only limited success. Arguably, this was due to these artificial systems being based on too simplistic a view of the immune system. We present here a second generation artificial immune system for process anomaly detection. It improves on earlier systems by having different artificial cell types that process information. Following detailed information about how to build such second generation systems, we find that communication between cells types is key to performance. Through realistic testing and validation we show that second generation artificial immune systems are capable of anomaly detection beyond generic system policies. The paper concludes with a discussion and outline of the next steps in this exciting area of computer security.

## INTRODUCTION

The work discussed here is motivated by a broad interest in biologically-inspired approaches to computer security, particularly in immune-inspired approaches to intrusion detection. The first part of this chapter gives a brief overview of biologically-inspired computing and computer security, and introduces the field of artificial immune systems. We have developed an immune-inspired process anomaly detection system. Process anomaly detection is an important technique in computer security for detecting a range of attacks, and the second part of this chapter introduces and reviews current approaches to process anomaly detection, relating our work to other work in this area. The third section of this chapter introduces our own efforts to develop a prototype immune-inspired realtime process anomaly detection system. However, our interests are also wider, and address issues concerning how artificial immune systems are modelled and implemented in general. We have implemented a system, `libtissue`, in which immune-inspired algorithms can be developed and tested on real-world problems. The design and implementation of this system is briefly reviewed. The final part of this chapter presents and discusses the results of validation tests using `libtissue`. A number of datasets containing system call and signal information were generated and a simple algorithm was implemented to test the `libtissue` system. The behaviour of the algorithm is analysed and it is shown how the `libtissue` system can be used to build immune-inspired algorithms that detect anomalies in process behaviour.

## BIOLOGICALLY-INSPIRED APPROACHES

Biological approaches to computer security are appealing for a number of reasons. Williamson (2002), discusses some of these reasons and their impact on the design of computer security systems. Biological organisms have developed many novel, parsimonious and effective protection mechanisms. As computer systems and networks become more complex traditional approaches are often ineffective and suffer from problems such as scalability, and biologically systems are important sources of inspiration when designing new approaches. The short position paper of Morel (2002) discusses the general design of cyber-security systems that provides a large distributed computer network with a high degree of survivability. He proposes that a cyber-security system emulates the architecture of the biological immune system. As in this chapter, the innate immune system is considered as central to the immune response, processing information and controlling the adaptive immune system. An effective cyber-security system should emulate key features, most importantly distributed control, of the biological system, it should provide multiple information gathering mechanisms, and it should coevolve with the threat.

In another interesting position paper Williams (1996) explores the similarities between people's health and the security of complex computer systems. Humans are composed of distinct but tightly integrated multilayer systems, have external interfaces which can receive a wide range of input and which carefully balance security and functionality, and have internal interfaces with protection mechanisms. They are not born with many of their defenses but learn to protect themselves against recurring threats such as viruses, and are able to identify and develop defenses for new threats. The body is able to detect conditions that are likely to lead to injury. It is surrounded by a skin which, if damaged, leads to further response. Williams suggests that computer systems also need to have virtual skins with a similar functionality. He highlights the importance of the balance between functionality, security and flexibility. Humans, as with computer systems, live a complex environment where conditions change over time. Both computer and biological systems are very sensitive to the input they receive. Biological systems check and filter input at many levels and he suggests security systems need to do the same. He also emphasises the impossibility of accurate measurement of health in humans, which is reflected in the difficulty of measuring the security of computer systems. His general view is that the computer security industry is becoming as specialised as the healthcare industry, with security engineers akin to doctors.

Our interest is in immune-inspired approaches to intrusion detection. The field of artificial immune systems began in the early 1990s with a number of independent groups conducting research which used the biological immune system as inspiration for solutions to problems in other domains. There are several general reviews of artificial immune system research (Dasgupta, 2006, Hart and Timmis, 2005), and a number of books including Dasgupta (1999) and de Castro and Timmis (2002) covering the field. Large bibliographies have been collated by Dasgupta and Azeem (2006) (over 600 journal and conference papers) and an annual international conference has been held since 2002 (Proceedings of the International Conference on Artificial Immune Systems, 2002-2007). Specifically of relevance to this chapter is the review of artificial immune system approaches to intrusion detection by Aickelin et al. (2004).

Intrusion detection systems are software systems designed to identify and prevent the misuse of computer networks and systems. Still a relatively young field, first discussed by James Anderson in his seminal 1980 paper (Anderson, 1980) and with the first working system described in Dorothy Denning's 1987 paper (Denning, 1987), intrusion detection still faces many unresolved research issues. Many intrusion detection systems have been developed, representative samples of which are reviewed in Kemmerer and

Vigna (2002) and Venter and Eloff (2003). Several excellent review papers (Axelsson, 2000, Bace and Mell, 2001) and books (Northcutt and Novak, 2003) on intrusion detection approaches have also been published. There are a number of different ways to classify intrusion detection systems (Axelsson, 2000). In their paper, Jansen and Karygiannis (1999) discuss the shortcomings of current intrusion detection system technology and the advantages of and approaches to applying mobile agents to intrusion detection and response. They highlight the issue of false positives as the primary problem facing the intrusion detection system community, and this is one of the key issues which this chapter seeks to address, particularly in terms of the detection of novel attacks.

## PROCESS ANOMALY DETECTION

In the classic paper "An Evening with Berferd in which a Cracker is Lured, Endured, and Studied" (Cheswick, 1992), Cheswick describes the activities of a cracker who is allowed to gain access to a monitored machine. Other more recent publications (Mitnick and Simon, 2005) which have deconstructed real-world attacks have painted a similar picture. Often, the initial goal of an attack is to gain administrator privileges or "get root" on a machine and so give the attacker free reign on the system. If the attacker does not have an account on the system then they may try to exploit a vulnerability in a network service running on the target remote machine to gain access. This is termed a remote-to-local attack. Buffer overflow exploits are often used to subvert remote services to execute code the attacker supplies and, for example, open a remote command shell on the target machine. Sometimes, the attacked service will already be running with administrator privileges, in which case the initial attack is complete. Otherwise, the attacker will have access to the machine at the same privilege level as the attacked service is running at. In this case, or if the attacker already has a local user account on the target machine, they will need to perform a privilege escalation attack, called a user-to-root attack. Often, this will involve attacking a privileged program, such as a program running with administrator privileges, and, once again, subverting its execution to create a command shell with administrator privileges. After the initial goal of unrestricted access is achieved, the attacker may install rootkits to hide their presence and facilitate later access. Data can be copied to and from the machine, remote services such as filesharing daemons can be started, and, in the case of worms, this may all be done automatically without human intervention.

Process anomaly detection systems are designed to detect and prevent the subversion of processes necessary in such remote-to-local and user-to-root attacks. A number of host-based intrusion detection systems have been built around monitoring running processes to detect intrusions, and are discussed in detail in the next section. In general, these intrusion detection systems collect information about a running process from a variety of sources, including from log files created by the process, or from other information gathered by the operating system. The general idea is that by observing what the process is currently doing e.g. by looking at its log files, we can tell whether the process is behaving normally or has been subverted by an attack. While log files are an obvious starting point for such systems, and are still an important component in a holistic security approach, it is fairly easy to execute attacks which do not cause any logging to take place, and so evade detection. Because of this, there has been a substantial amount of research into other data sources, usually collected by the operating system. Of these, system calls have been the most favoured approach. This section begins with a brief background on system calls and then reviews current system call-based approaches to process anomaly detection.

## Processes and System Calls

A process is a running instance of a program. On modern multitasking operating systems many processes are effectively running simultaneously. For example, a server may be running a web server, email servers and a number of other services. A single program executable, when run, may create several child processes by forking (fork, 2007) or threading (pthreads, 2007). For example, web servers typically start child processes to handle individual connections once they have been received. The process which created the child process is called the parent process. Child processes themselves may create children, sometimes generating a complex process tree derived from a single parent process node created when the executable is first run. The operating system is responsible for managing the execution of running processes, and associates a number, called a process identifier, with each process. This number uniquely identifies a process. Essentially, the operating system initialises a counter and assigns its value to a new process, and then increments the counter. When a process is started, the operating system associates other metadata with it too, such as the process identifier of the parent process that created it, and the user who started the process. The process is also allocated resources by the operating system. These resources include memory, which stores the executable code and data, and file descriptors, which identify files or network sockets which belong to the process.

Systemcalls (syscalls) are a low-level mechanism by which processes request system resources such as peripheral I/O or memory allocation from an operating system. As a process runs it cannot usually directly access memory or hardware devices. Instead, the operating system manages these resources and provides a set of functions, called syscalls, which processes can call to access these resources. On modern Linux systems there are around 300 syscalls, accessed via wrapper functions in the libc library. Some of the more common syscalls are summarised in Table 1.1. At an assembly code level, when a process wants to make a syscall it will load the syscall number into the EAX register, and syscall arguments into registers such as EBX, ECX or EDX. The process will then raise the 0x80 interrupt. This causes the process to halt execution and the operating system to execute the requested syscall. Once the syscall has been executed, the operating system places a return value in EAX and returns execution to the process. Operating systems other than Linux differ slightly in these details, for example BSD puts the syscall number in EAX and pushes the arguments onto the stack (Bovet and Cesati, 2002, syscalls, 2007). Higher-level languages provide library calls which wrap the syscall in easier-to-use functions such as printf.

Table 1.1: Common system calls (syscalls).

| number | name | description |
|---|---|---|
| 1 | exit | terminate process execution |
| 2 | fork | fork a child process |
| 3 | read | read data from a file or socket |
| 4 | write | write data to a file or socket |
| 5 | open | open a file or socket |
| 6 | close | close a file or socket |
| 37 | kill | send a kill signal |
| 90 | old_mmap | map memory |
| 91 | munmap | unmap memory |
| 301 | socket | create a socket |
| 303 | connect | connect a socket |

Syscalls are a much more powerful data source for detecting attacks than log file entries. All of a monitored application's interactions with the network, filesystem, memory, and other hardware devices can be monitored. Most attacks which subvert the execution of the monitored application will probably have to access some of these resources, and so will have to make a number of syscalls. In other words, it is much harder to evade a syscall-based intrusion detection system. However, monitoring syscalls is more complex and costly than reading data from a log file, and usually involves placing hooks or stubs in the operating system, or wrapping the monitored process in a sandbox-like system. This increases the runtime of the monitored process, since for each syscall the monitor will spend at least a few clock ticks pushing the data it has collected to a storage buffer. Syscall interposition systems, which, in addition to passively gathering syscall information, also decide whether to permit or deny the syscall, can add additional runtime overheads. Also, processes can generate hundreds of syscalls a second, making the data load significantly higher. Other factors also need to be taken into account when implementing a syscall monitoring or interposition system. Incorrect replication of operating system state or other race conditions may allow syscall monitoring to be circumvented. These factors are addressed in detail by Garfinkel (2003).

## Current Approaches

The `systrace` system of Provos (2003) is a syscall-based confinement and intrusion detection system for Linux, BSD and OSX systems. A kernel patch inserts various hooks into the kernel to intercept syscalls from the monitored process. The user specifies a syscall policy which is a whitelist of permitted syscalls and arguments. The monitored process is wrapped by a user-space program which compares any syscalls a process tries to make (captured by the kernel hooks) with this policy, and only allows the process to execute syscalls which are present on the whitelist. Execution of the monitored process is halted while this decision is made, which, along with other factors such as the switch from kernel- to user-space, adds an overhead to the monitored process. However, due in part to the simplicity of the decision-making algorithm (a list search on the policy file), as well as a good balance of kernel- versus user-space, this performance impact on average is minimal, and `systrace` has been used to monitor processes in production environments. As an intrusion detection system, `systrace` can be run to either automatically deny and log all syscall attempts not permitted by the policy, or to graphically prompt a user as to whether to permit or deny the syscall. The latter mode can be used to add syscalls to the policy, adjusting it before using it in automatic mode. Initial policies for a process are obtained by using templates or by running `systrace` in automatic policy generation mode. In this mode, the monitored process is run under normal usage conditions and permit entries are created in the policy file for all the syscalls made by the process. The policy specification allows some matching of syscall arguments as well as syscall numbers.

Gao et al. (2004) introduce a new model of syscall behaviour called an execution graph. An execution graph is a model that accepts approximately the same syscall sequences as would a model built on a control flow graph. However, the execution graph is constructed from syscalls gathered during normal execution, as opposed to a control flow graph which is derived from static analysis. In addition to the syscall number, stack return addresses are also gathered and used in construction of the execution graph. The authors also introduce a course-grain classification of syscall-based intrusion detection systems into white-box, black-box and gray-box approaches. Black-box systems build their models from a sample of normal execution using only syscall number and argument information. Gray-box approaches, as with black boxes, build their models from a sample of normal execution but, as well as using syscall information, also use additional runtime information. White-box approaches do not use samples of normal execution, but instead use static analysis techniques to derive their models. A prototype gray-box anomaly detection system using execution graphs is introduced by the authors, and they compare this approach to other systems and discuss possible evasion strategies in Gao et al. (2004).

Sekar et al. (2001) implement a realtime intrusion detection system which uses finite state automata to capture short and long term temporal relationships between syscalls. One advantage of using finite state automata is that there is no limit to the length of the syscall sequence. Yeung et al. (2003) describe an intrusion detection system which uses a discrete hidden Markov model trained using the Baum-Welch re-estimation algorithm to detect anomalous sequences of syscalls. Lee (2000) explores a similar Markov chain model. Krugel et al. (2003) describe a realtime intrusion detection system implemented using Snare under Linux. Using syscall audit logs gathered by Snare, their system automatically detects anomalies in syscall arguments. They explore a number of statistical models which are learnt from observed normal usage. Endler (1998) presents an offline intrusion detection system which examines BSM audit data. They combine a multi-layer perceptron neural network which detects anomalies in syscall sequences with a histogram classifier which calculates the statistical likelihood of a syscall. Lee and Xiang (2001) evaluate the performance of syscall-based anomaly detection models built on information-theoretic measures such as entropy and information cost, and also use these models to automatically calculate parameter settings for other models.

Forrest, Hofmeyr, Somayaji and other researchers at the University of New Mexico have developed several immune-inspired learning-based approaches. In Forrest et al. (1997) and Hofmeyr and Forrest (2000) a realtime system is evaluated which detects anomalous processes by analysing sequences of syscalls. Syscalls generated by an application are grouped together into sequences, in this case sequences of six consecutive syscalls. This choice of sequence length is discussed in Tan and Maxion (2003). A database of normal sequences is constructed and stored as a tree during training. Sequences of syscalls are then compared to this database using a Hamming distance metric, and a sufficient number of mismatches generates an alert. No user-definable parameters are necessary, and the mismatch threshold is automatically derived from the training data. Similar approaches have also been applied by this group to network intrusion detection (Balthrop et al., 2002, Hofmeyr and Forrest, 2000). Somayaji (2002) develops the immune-inspired pH intrusion prevention system which detects and actively responds to changes in program behaviour in realtime. As with the method just described, sequences of syscalls are gathered for all processes running on a host and compared to a normal database using a similar immune-inspired model. However, if an anomaly is detected, execution of the process that produced the syscalls will be delayed for a period of time. This method of response, as opposed to more malign responses such as killing a process, is more benign in that if the system makes a mistake and delays a process which is behaving normally, this may not have a perceptible impact from the perspective of the user. The idea of process homeostasis, with pH maintaining a host machine within certain operational limits, is introduced. This approach was effective at automatically preventing a number of attacks.

## THE `libtissue` SYSTEM

The broader aim of the research presented here is to build a software system which allows researchers to implement and analyse novel artificial immune system algorithms and apply them to real-world problems. We have implemented a prototype of such a system, called `libtissue`, which is being used by ourselves and other researchers (Greensmith et al., 2006a, 2006b, Twycross and Aickelin, 2006) to build and evaluate novel immune-inspired algorithms for process anomaly detection. This section briefly reviews the design and implementation of the `libtissue` system, more detail of which can be found in (Twycross and Aickelin, 2006).

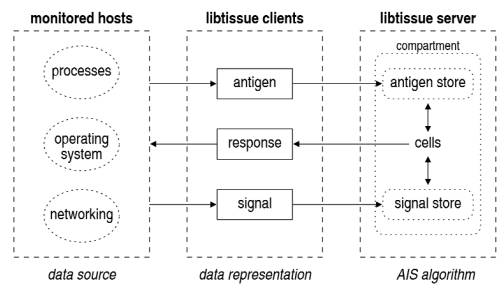

Figure 1.1: The architecture of `libtissue`. Hosts are monitored by `libtissue` antigen and signal clients, which in turn provide input data to the artificial immune system algorithm, run on a `libtissue` server. Algorithms are able to change the state of the monitored hosts through response clients.

`libtissue` has a client/server architecture pictured in Figure 1.1. An artificial immune system algorithm is implemented as part of a `libtissue` server, and `libtissue` clients provide input data to the algorithm and response mechanisms which change the state of the monitored system. This client/server architecture separates data collection by the `libtissue` clients from data processing by the `libtissue` servers and allows for relatively easy extensibility and testing of algorithms on new data sources.

`libtissue` was coded in C as a Linux shared library with client and server APIs, allowing new antigen and signal sources to be easily added to `libtissue` servers from a programmatic perspective. Because `libtissue` is implemented as a library, algorithms can be compiled and run on other researchers' machines with no modification. Client/server communication is socket-based, allowing clients and servers to potentially run on separate machines, for example a signal or antigen client may in fact be a remote network monitor.

Artificial immune system algorithms are implemented within a `libtissue` server as multiagent systems of cells. Cells exist within an environment, called a tissue compartment, along with other cells, antigen and signals. The problem to which the algorithm is being applied is represented by `libtissue` as antigen and signals. Cells express various repertoires of receptors and producers which allow them to interact with antigen and control other cells through signalling networks. `libtissue` allows data on implemented algorithms to be collected and logged, allowing for experimental analysis of the system. A `libtissue` server is in fact several threaded processes running asynchronously. An initialisation routine is first called which creates a tissue compartment based on user-supplied parameters. During initialisation a thread is also started to handle connections between the server and `libtissue` clients, and this thread itself starts a separate thread for each connected client. After initialisation, cells, the characteristics of which are specified by the user, are created and initialised, and the tissue compartment populated with these cells. Cells in the tissue compartment then cycle and input data is provided by connected `libtissue` clients.

`libtissue` clients are of three types: antigen, signal and response. Antigen clients collect and transform data into antigen which are forwarded to a `libtissue` server. Currently, a `systrace` antigen client has been implemented which collects process syscalls using `systrace` (systrace homepage, 2007). Signal clients monitor system behaviour and provide an artificial immune system running on the tissue server with input signals. A process monitor signal client, which monitors a process and its children and records statistics such as CPU and memory usage, and a network signal client, which monitors network interface statistics such as bytes per second, have been implemented. Two response clients have been implemented, one which simply logs an alert, and another which allows an active response through the modification of a `systrace` syscall policy. All of these clients are designed to be used in realtime experiments and for data collection for offline experiments with `tcreplay`.

The implementation is designed to allow varied artificial immune system algorithms to be evaluated on real-world, realtime systems and problems. When testing intrusion detection systems it is common to use preexisting datasets such as the Lincoln Labs dataset (Lincoln Labs DARPA Intrusion Detection Evaluation datasets, 2007). However, the project `libtissue` has been built for is focused on combining measurements from a number of different concurrent data sources. Preexisting datasets which contain all the necessary sources are not available. Therefore, to facilitate experimentation, a `libtissue` replay client, called `tcreplay`, was also implemented. This client reads in log files gathered from previous realtime runs of antigen and signal clients. It also has the facility to read logfiles generated by `strace` (strace homepage, 2007) as an optional source of antigen in place of the `systrace` client. It then sends the information in these logs to a `libtissue` server. Variable replay rates are available, allowing data collected from a realtime session to be used to perform many experiments quickly. Having such a replay facility is important in terms of reproducibility of experiments. In reality, the majority of experimental runs are scripts which take data and parameter files as input and run a tissue server and `tcreplay` client.

## VALIDATION OF APPROACH

We wanted to verify that useful algorithms could be implemented and applied to a real-world problem. This section reviews the details of this validation. It discusses how data on a process anomaly detection problem was generated. It then presents a simple anomaly detection algorithm which we have implemented to test the `libtissue` system. Results from an analysis of the behaviour and performance of the `libtissue` system and the algorithm are then presented. Lastly, an example of how this algorithm and the `libtissue` system can be used to detect anomalies in process behaviour is given.

## Dataset Generation

In order to gather data for the process anomaly detection problem, a small experimental network with three hosts was set up. One host, the target, runs software, in this case a Redhat 6.2 server, with a number of vulnerabilities. The other two hosts act as clients which interact with the target machine, either attempting to exploit its vulnerabilities or simulating normal usage. Because the experimental network contains vulnerable hosts, access between it and the public campus network is tightly controlled. While minimal, this setup permits realistic network-based attack and normal usage scenarios to be played out. Physically, the network exists on a single Debian Linux host running two VMware guest operating systems. The host and guests are connected via a virtual VMware host-only network. This setup was chosen as it allows for relatively fast configuration and restoration of the experimental network when compared with one in which each host is a physically separate machine connected via the standard network infrastructure of switches and so on. Redhat 6.2 was chosen because the default installation installs a number of programs with vulnerabilities (Redhat Linux 6.2 Security Advisories, 2002) and because many well-documented exploits are available for these vulnerabilities. Tests were carried out with the `rpc.statd` daemon (rpc.statd, 2007), which provides a status monitoring service to other NFS clients and servers. The default version of `rpc.statd` shipped with Redhat 6.2 has a format string vulnerability which allows a remote user to execute arbitrary code with root privileges on the server (Multiple Linux Vendor rpc.statd Remote Format String Vulnerability, 2000). An exploit, `statdx2` (Bugtraq: statdx2 - Linux rpc.statd revisited, 2002), has been released which levers this vulnerability and, by default, injects shellcode which causes a remote root shell to be opened on the attacker's machine, allowing unrestricted access to the server. This vulnerability has also been used in automated attacks such as the Ramen worm.

In order to collect the data, that is process syscall information and appropriate context signals, the target system was instrumented. The Redhat nfslock init script was modified to start `rpc.statd` wrapped by `strace` (strace homepage, 2007), which logged all the syscalls made by `rpc.statd` and its children. At the same time, a specially written application called `process_monitor` was started which monitors a process and all of its child processes. At regular intervals, one tenth of a second in this case, it takes a snapshot of the process table which it then traverses, recording the process identifiers of all the processes which are children of the monitored process. The monitor then logs the current name of the monitored process, the total number of children including itself owned by the process, the total CPU usage of the process and its children, and the total memory usage of the process and its children. Pairs of `strace` and `process_monitor` logs were collected on the instrumented target machine while `rpc.statd` was utilised in a number of different scenarios. These logs were then parsed to form a single `tcreplay` logfile for each of the scenarios. An antigen entry in the `tcreplay` log was created for every syscall recorded in the `strace` log. A signal entry was created for each recording of CPU usage in the `process_monitor` log. While the `strace` log actually contains much more information, the use of just the syscall number is more than sufficient for testing the example algorithm described in the next section. It would be expected that a more complex algorithm would require additional complexity in both the antigen and range of signals it is provided with, such as the addition of information about syscall arguments, sequences of syscalls, or instruction pointer addresses.

The monitored scenarios are divided into three groups based on whether the type of interaction with the `rpc.statd` server is a successful attack, a failed attack, or normal usage. Statistics for the datasets are given in Table 1.2. All the datasets followed a similar pattern. The data was generally very bursty in terms of syscalls per second, with relatively long periods of no syscalls punctuated by bursts of up to 1102 syscalls per second (success1). All datasets contain an initial one second burst of 405 syscalls executed by `rpc.statd` during normal startup. Syscalls generated by `rpc.statd` at shutdown, a burst of between 17 and 29 syscalls caused by typing halt on the server, are also present in the normal and failure datasets. They are not present in the success datasets as the `rpc.statd` process is replaced by a shell process during the exploit and so not able to go through normal shutdown. In both successful attacks there are three bursts of between 98 and 1102 syscalls. The user interaction on the resulting remote shell (typing `exit`) creates 5 syscalls. The unsuccessful attacks produced a single burst of 96 and 62 syscalls (failure1 and failure2 respectively). The actions of the NFS client in normal2 result in a single burst of 16 syscalls.

Table 1.2: Statistics for the six datasets gathered.

| dataset | total time | total antigen | max antigen rate |
|---|---|---|---|
| success1 | 55 | 1739 | 1102 |
| success2 | 36 | 1743 | 790 |
| failure1 | 54 | 518 | 405 |
| failure2 | 68 | 495 | 405 |
| normal1 | 38 | 434 | 405 |
| normal2 | 104 | 450 | 405 |

# The `twocell` Algorithm

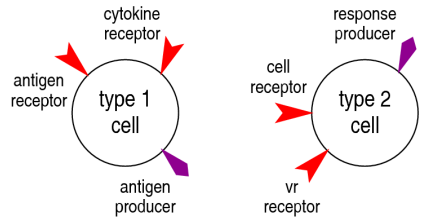

Figure 1.2: The two different cell types implemented in `twocell`.

The cells in `twocell`, shown in Figure 1.2, are of two types, labelled Type 1 and Type 2, and each type has different receptor and producer repertories, as well as different cell cycle callbacks. Type 1 cells are designed to emulate two key characteristics of biological APCs: antigen and signal processing. In order to process antigen, each Type 1 cell is equipped with a number of antigen receptors and producers. A cytokine receptor allows Type 1 cells to respond to the value of a signal in the tissue compartment. Type 2 cells emulate three of the characteristics of biological T cells: cellular binding, antigen matching, and antigen response. Each Type 2 cell has a number of cell receptors specific for Type 1 cells, VR (variable-region) receptors to match antigen, and a response producer which is triggered when antigen is matched. Type 2 cells also maintain one internal cytokine, an integer which is incremented every time a match between an antigen producer and VR receptor occurs. If the value of this cytokine is still zero, that is no match has occured, after a certain number of cycles, set by the cell_lifespan parameter, then the values of all of the VR receptor locks on the cell are randomised.

A tissue compartment is created and populated with a number of Type 1 and 2 cells. Antigen and signals in the compartment are set by `libtissue` clients based on the syscalls a process is making and its CPU usage. Type 1 and 2 cells have different cell cycle callbacks. Type 1 cells ingest antigen through their antigen receptors and present it on their antigen producers. The period for which the antigen is presented is determined by a signal read by a cytokine receptor on these cells, and so can be made dependant upon CPU usage. Type 2 cells attempt to bind with Type 1 cells via their cell receptors. If bound, VR receptors on these cells interact with antigen producers on the bound Type 1 cell. If an exact match between a VR receptor lock and antigen producer key occurs, the response producer on Type 2 cells produces a response, in this case a log entry containing the value of the matched receptor.

## System Dynamics

In experiments it is important to have a baseline with which to compare algorithmic performance. In terms of syscall policies such a baseline can be generated, and is here termed a naive policy. A naive syscall policy is generated for a process, such as `rpc.statd`, by recording the syscalls it makes under normal usage, as in the normal1 and normal2 datasets. A permit policy statement is then created for all syscalls seen in the datasets. This baseline is not too unrealistic when compared to how current systems such as `systrace` automatically generate a policy. Similarly to a naive policy, one way in which `twocell` can be used to generate a syscall policy is by running it with normal usage data during a training phase. During the run, responses made by Type 2 cells are recorded. At the end of each run, a syscall policy is created by allowing only those syscalls responded to, and denying all others. Since interactions in `libtissue` are stochastic, looking at the average results over a number of runs helps to understand the behaviour of implemented algorithms. A script was written to start the `twocell` server and then after 10 seconds start the `tcreplay` client and replay a dataset in realtime. `twocell` was allowed to continue running for a further minute after replay had finished. This process was repeated 20 times for both the normal1 and normal2 datasets, yielding 40 individual syscall policies. A single average `twocell` policy was then generated by allowing all syscalls which were permitted in any of the 40 individual policies. It was found that all of the 38 syscalls that were permitted in the naive policy were also permitted in the average policy.

Table 1.3: The syscall policy generated by `twocell` for the normal2 dataset and the frequency of response for each syscall.

| syscall | frequency |
|---|---|
| gettimeofday(78) | 1 |
| listen(304) | 1 |
| send(309) | 1 |
| select(142) | 2 |
| poll(168) | 3 |
| recvfrom(312) | 8 |
| fcntl(55) | 9 |
| fstat(108) | 9 |
| open(5) | 22 |
| close(6) | 34 |

In order to examine more closely how `twocell` responds, a single run of the `twocell` algorithm was observed. Following the same general procedure as the previous experiment, `twocell` was run once with the normal2 dataset. The resulting policy is shown in Table 1.3, along with the frequencies with which the permitted syscalls were responded to. During the run, the time at which a Type 2 cell produced a response to a particular syscall was also recorded, and the rate at which these responses occured was calculated. This clearly showed a correlation between the rate of incoming syscalls and the rate of responses produced by Type 2 cells. Cells initially do not produce any response until syscalls occur, and then produce a burst of responses for a relatively short period before settling down to an unresponsive state once again. This is to be expected, as antigen (syscalls) enter and are passed through `twocell` until their eventual destruction after being presented on Type 1 cell antigen producers.

## Classification Accuracy

An example is now given of how the classification accuracy and error of a `libtissue` algorithm can be evaluated. In terms of syscall policies, a particular policy can be considered successful in relation to the number of normal syscalls it permits versus the number of attack syscalls it denies. The naive policy and average `twocell` policy generated from datasets normal1 and normal2 in the experiment above were evaluated in such a way. The number of syscalls both policies permitted and denied when applied to the four datasets in the attack and failed attack groups was recorded. Syscalls within these groups were labelled as either generated by an attack or by normal usage. For each dataset, Table 1.4 shows the percentages of attack and normal syscalls in the dataset, together with the percentage of syscalls permitted by the naive and `twocell` policies. From the results, the tendency of the naive policy was to permit the vast majority of syscalls, whether attack related or not. The `twocell`-generated policy behaved much more selectively, denying a slightly larger proportion of syscalls in the success1 and success2 datasets than it permitted. For the failure1 and failure2 dataset the converse was true.

Table 1.4: Comparison of the performance of a naive policy and a `twocell` policy generated from the normal2 dataset.

| dataset | success1 | success2 | failure1 | failure2 |
|---|---|---|---|---|
| normal syscalls | 23% | 23% | 81% | 87% |
| attack syscalls | 76% | 76% | 18% | 12% |
| naive permit | 90% | 90% | 99% | 99% |
| naive deny | 9% | 9% | 0% | 0% |
| twocell permit | 47% | 47% | 69% | 68% |
| twocell deny | 52% | 52% | 30% | 31% |

## Discussion

The dataset, algorithm and experiments presented in this section show how a novel algorithm has been developed and applied using the `libtissue` system. Runs used on average around 1%, and never more than 3%, of the available CPU resources, showing that it is computationally viable to process realtime data using our approach. The collection and analysis of the `rpc.statd` data has brought to light the potential usefulness of a number of novel data sources which can be use in conjuction with syscall information. The experiments we conducted compared an algorithm, `twocell`, implemented with `libtissue`, to a baseline standard approach, and showed how the agents in `twocell` responded in different ways to normal and attack sessions. By measuring the response of the agents, we use our algorithm to classify sessions as normal or attack. This experiment showed that the performance `twocell` is at least comparable to current approaches.

More widely, the validation experiments with the `twocell` algorithm and the `rpc.statd` dataset show the feasibility of using `libtissue` to implement artificial immune systems as multiagent systems and apply them to real-world problems. The

`twocell` algorithm has also provided a necessary stepping-stone on the route to developing more complex algorithms. We are preparing to publish results of experiments with an algorithm which is able to detect a number of novel attacks with a low false-positive rate. To evaluate this and other algorithms we have created a second dataset which contains a wider range of normal and attack usage that the `rpc.statd` dataset. The new dataset, which was created by monitoring a `wuftpd` FTP server, contains syscalls and 13 different signals including CPU usage, memory usage, and socket and file usage statistics.

In order to generate realistic normal usage of the `wuftpd` server, we recreated normal sessions seen on a production network on an instrumented experimental network much like the setup for the `rpc.statd` dataset. Data on real FTP client-server interaction can be readily obtained from network packet traces collected by network-based monitors. Such packet traces are technically fairly easy to gather but, more importantly, traces are also already available publically, removing the need to gather this data altogether. Use of public datasets also contributes to the reproducibility of experiments. By reassembling network packets transmitted between client and server a sufficiently complete record of an FTP session can be reproduced. The dataset used (LBNL-FTP-PKT dataset, 2003) contains all incoming anonymous FTP connections to public FTP servers at the Lawrence Berkeley National Laboratory over a ten-day period and is available from the Internet Traffic Archive (Internet Traffic Archive, 2007). The traces contain connections between 320 distinct FTP servers and 5832 distinct clients and provide a rich source of normal usage sessions, and we initially used the traces for one FTP server over two days.

## CONCLUSIONS

In this chapter we have given a overview of biologically-inspired approaches to computer security, in particular immune-inspired approaches. We then discussed in detail an intrusion detection problem, process anomaly detection, and reviewed current research in this area. A system, `libtissue`, which we have built for implementing immune-inspired algorithms was then detailed, and the results of validation experiments using an artificial immune system implemented with `libtissue` and applied to process anomaly detection were presented and discussed.

# KEYWORDS

Artificial Immune System: A relatively new class of meta-heuristics that mimics aspects of the human immune system to solve computational problems. This method has shown particular promise for anomaly detection. Previous artificial immune systems have shown some similarities with evolutionary computation. This is because they focus on the adaptive immune system. More recent approaches have combined this with aspects of the innate immune system to create a second generation of artificial immune systems.

Adaptive Immune System: Central components of the adaptive immune system are T cells and B cells. The overall functionality of the adaptive immune system is to try and eliminate threats through antibodies, which have to be produced such that they match antigen. This is achieved in an evolutionary-like manner, with better and better matches being produced over a short period of time. The adaptive system remembers past threats and hence has the capability of responding faster to future similar events.

Innate Immune System: Central components of the innate immune system are antigen presenting cells and in particular dendritic cells. Until recently, the innate system was viewed as less important than the adaptive system and its main function was seen as an information pre-processing unit. However, the latest immunological research shows that it is the innate system that actually controls the adaptive system. Above all, dendritic cells seem to be the key decision makers.

T Cells: Created in the thymus (hence the 'T'), these cells come in different subtypes. Cytotoxic T cells directly destroy infected cells. T helper cells are essential to activate other cells, e.g. B cells. T reg cells suppress inappropriate responses.

Dendritic Cells: These belong to the class of antigen presenting cells. During their life, dendritic cells ingest antigen and redisplay it on their surface. In addition, dendritic cells mature differently depending on the context signals they are exposed to. Using these two mechanisms, these cells differentiate between dangerous and non-dangerous material and then activate T cells.

Process Anomaly Detection: A method of detecting intrusions on computer systems. The aim is to detect misbehaving processes, as this could be a sign of an intrusions. The detection is based on syscalls, i.e. activities by the processes, and context signals, e.g. CPU load, memory usage or network activity.